\documentclass[runningheads]{llncs}
\usepackage{graphicx}
\begin{document}
\title{LLM2KB: Constructing Knowledge Bases using instruction tuned context aware Large Language Models}
\titlerunning{LLM2KB}
%

\author{Anmol Nayak$^*$ \and
Hari Prasad Timmapathini$^*$}
\authorrunning{A. Nayak and H. P. Timmapathini}
%
\institute{ARiSE Labs at Bosch, Bangalore, India
\email{\{Anmol.Nayak,HariPrasad.Timmapathini\}@in.bosch.com}}

\maketitle              
\def\thefootnote{*}\footnotetext{Both authors contributed equally to this work}
\def\thefootnote{+}\footnotetext{Submission: https://github.com/anmoln94/Team\_LLM2KB\_LM-KBC-2023}

\begin{abstract}
The advent of Large Language Models (LLM) has revolutionized the field of natural language processing, enabling significant progress in various applications. One key area of interest is the construction of Knowledge Bases (KB) using these powerful models. Knowledge bases serve as repositories of structured information, facilitating information retrieval and inference tasks. Our paper proposes LLM2KB, a system for constructing knowledge bases using large language models, with a focus on the Llama 2 architecture and the Wikipedia dataset. We perform parameter efficient instruction tuning for Llama-2-13b-chat and StableBeluga-13B by training small injection models that have only $\approx$0.05 \% of the parameters of the base models using the Low-Rank Adaptation (LoRA) technique. These injection models have been trained with prompts that are engineered to utilize Wikipedia page contexts of subject entities fetched using a Dense Passage Retrieval (DPR) algorithm, to answer relevant object entities for a given subject entity and relation. Our best performing model achieved an average F1 score of 0.6185 across 21 relations in the LM-KBC challenge held at the ISWC 2023 conference.

\end{abstract}

\section{Introduction}
The rapid advancements in natural language processing (NLP) have propelled the development of large language models, revolutionizing the way machines understand and generate human language. One of the pivotal applications of these sophisticated models lies in the construction of knowledge bases, which serve as repositories of structured information essential for a multitude of NLP tasks, including information retrieval, question answering, and knowledge inference.

Knowledge bases hold immense potential for enhancing machine understanding of the world, but constructing them manually is a laborious and time-consuming process. However, the emergence of large language models such as GPT-4~\cite{gpt}, Llama 2~\cite{llama2} Stable Beluga 2~\cite{beluga}, has opened new possibilities for knowledge base construction. These models possess extensive linguistic knowledge and factual knowledge that can be leveraged for automated entity recognition, relation extraction, and knowledge representation.

One of the prominent sources for constructing knowledge bases is the vast repository of human-curated information available on Wikipedia. This publicly accessible dataset contains an extensive wealth of knowledge on diverse topics, making it an ideal resource for building comprehensive knowledge bases. Integrating Wikipedia data into knowledge bases allows for a wider coverage and a well-rounded understanding of various domains.

Nevertheless, the success of constructing knowledge bases using large language models hinges on the ability to fine-tune these models effectively. Traditional fine-tuning methods often suffer from scalability issues and demand an excessive amount of computational resources. However, recent advancements in parameter-efficient fine-tuning techniques, like LoRA (Low-Rank Adaptation)~\cite{lora}, have showcased promising results in reducing the complexity and computational requirements while preserving model performance. By efficiently fine-tuning large language models, researchers can unlock their true potential in knowledge base construction and empower them to comprehend and generate valuable information.

In this paper, we describe our system developed for Track 2 of the LM-KBC challenge at ISWC 2023~\cite{lmkbc2023}, which focuses on using language models of any size for knowledge base construction. Given an input subject-entity (s) and relation (r), the system attempts to predict all the correct disambiguated object-entities.

\section{Related Work}

The inquiry into the potential of language models (LM) in replacing or aiding the creation and curation of knowledge bases was initially posed by~\cite{petroni} and later explored by other researchers~\cite{lmandfor}. The LAMA dataset which probes relational knowledge in language models through masked language modeling tasks to complete cloze-style sentences was introduced in~\cite{petroni}.

While the initial study utilized manually designed prompts to probe the language model, subsequent research has demonstrated the advantages of automatically learning prompts. Various methods have emerged to mine prompts from large text corpora and select the most effective ones~\cite{knowwhatlmknows,inducbert}. Additionally, prompts can be directly learned through back-propagation~\cite{9,10}, showcasing how learned prompts can enhance the performance on LAMA tasks.

The performance of probing language models can be significantly improved through various approaches, such as directly learning continuous embeddings for prompts~\cite{11,12}, fine-tuning the LM on the training data~\cite{13}, or few-shot learning ~\cite{14}. The authors demonstrate that combining few-shot examples with learned prompts achieves the best probing results.

While probing language models has been extensively studied in the NLP community, the idea of utilizing language models to support knowledge graph curation has not received adequate attention~\cite{lmandfor}. Some works have demonstrated the combination of language models with knowledge bases to complete query results effectively~\cite{15}. Others have investigated how language models can be employed to identify errors in knowledge graphs~\cite{16} or explored using language models to weigh KG triples from ConceptNet for measuring semantic similarity. ~\cite{17} have showcased the utility of language models in entity typing by predicting entity classes using language model-based approaches.

Further, the LM-KBC challenge at ISWC 2022~\cite{lmkbc} produced interesting submissions on using LM for KB construction.~\cite{18} developed a system that performed task-specific pre-training of BERT, used prompt decomposition for generating candidate objects progressively, and employed adaptive thresholds for candidate selection. They utilized additional knowledge triples from Wikidata KB for BERT pre-training and experimented with cloze-style prompts, but found that masking nearby tokens of the object-entity did not improve performance. By mining prompts from Wikipedia and using an ensemble approach with averaged voting, they achieved final object-entity predictions. They also proposed sticky thresholds for candidate selection based on likelihood comparison.

~\cite{19} introduces the ProP system, which employs GPT-3~\cite{gpt3} under a few-shot setting for knowledge base (KB) construction. ProP uses various prompting techniques, including manual prompt creation and question-style prompts, to verify the accuracy of GPT-3 generated claims. They utilize context examples with specific properties, such as varying answer sets, subjects with empty answer sets, and question-answer pairs, to train the model effectively.

~\cite{20} utilized manual prompts, generated from three automated sources, and applied ensemble learning for final predictions. Descriptive information from Wikidata was used to create prompts through "middle-word", "dependency-based" and "paraphrasing-based" strategies. The BERT-large model was probed using these prompts, and the five most frequent and likely objects were selected from the ensemble. Before selecting the top-5 objects, the candidate list was post-processed by removing stopwords. The threshold for candidate selection was treated as a hyper-parameter, and its tuning was done separately for each relation on the train dataset.

~\cite{21} proposes manual prompts tailored for each relation to probe the BERT-large model, utilizing semantics and domain knowledge. They uniquely utilize word co-occurrences in context to design prompts and reason about the relationship between subjects and objects in questions to optimize prompt design for various relations, observing improvements through simple modifications like changing articles in the prompt for the "plays-instrument" relation.~\cite{22} conducts experiments with different manual prompts and candidate selection thresholds for each relation during BERT model probing. They also investigate the impact of selecting a larger number of options in the object list (100, 150, 180, or 200) on overall performance. Additionally, they create an ensemble of their manually crafted prompts to further enhance the probing results.

\section{The LM-KBC 2023 Challenge}
\label{lmkbc}

\subsection{Description} The LM-KBC 2023 challenge is centered around constructing disambiguated knowledge bases from language models based on given subjects and relations. Unlike existing probing benchmarks like LAMA~\cite{petroni}, LM-KBC 2023 does not assume any specific relation cardinalities, allowing subjects to have zero, one, or multiple object-entities. Submissions are required to not only rank predicted surface strings but also materialize disambiguated entities in the output. The evaluation process involves calculating precision and recall for each test instance compared to the ground-truth values. For every relation, the macro-averaged precision, recall and f1-score are computed. The final ranking of participating systems is determined based on the average F1-score across all 21 relations. Formally, the task involves predicting all correct object-entities ({o1, o2, ..., ok}) using LM probing given the input subject-entity (s) and relation (r).

The challenge comes with two tracks:
\begin{itemize}
    \item {Track 1}: A small-model track with low computational requirements (\textless 1 billion parameters) and usage of context is not allowed.
    \item {Track 2}: An open track, where participants can use any LM of their choice and usage of context is allowed.
\end{itemize}

\subsection{Dataset}

The LM-KBC dataset comprises both training and validation sets, encompassing 21 diverse relations. A test set containing only subject entities and relations was released towards the end of the challenge. The train, validation and test set each comprise of 1940 records. Each relation covers distinct subject-entities, and for each subject-relation pair, a comprehensive list of ground truth object-entities is provided. Each row in the dataset files includes the subject-entity ID, subject-entity name, a list of all possible object-entity IDs, a list of all possible object-entities, and the corresponding relation. The entity IDs correspond to the Wikidata ID. To perform instruction fine-tuning on the models, we consume the provided dataset and generate training samples in 2 ways:

\begin{itemize}
    \item Method 1: First, the Train set and Validation set are combined to produce a super set. For each record, we begin generating a separate instruction tuning dataset (see Section~\ref{prompts} for details on the Prompts) by slot filling Prompt 1 and 2. Thus, each record in the super set produces 2 samples for the instruction tuning dataset. Then, for each ObjectEntity in the record we slot fill Prompt 3, which additionally generates as many new samples for the instruction tuning dataset as there are object entities. The generated instruction tuning dataset is shuffled and we keep aside 1000 samples for validation and the 14310 samples are used for training.
    \item Method 2: This method also utilises Prompt 1,2 and 3 however the instruction tuning training dataset is generated only using the Train set, while the entire Validation set is used to generate the instruction tuning validation dataset. 7666 samples were used for training and 7644 samples were used for validation.
\end{itemize}

\begin{figure}
\includegraphics[width=\textwidth]{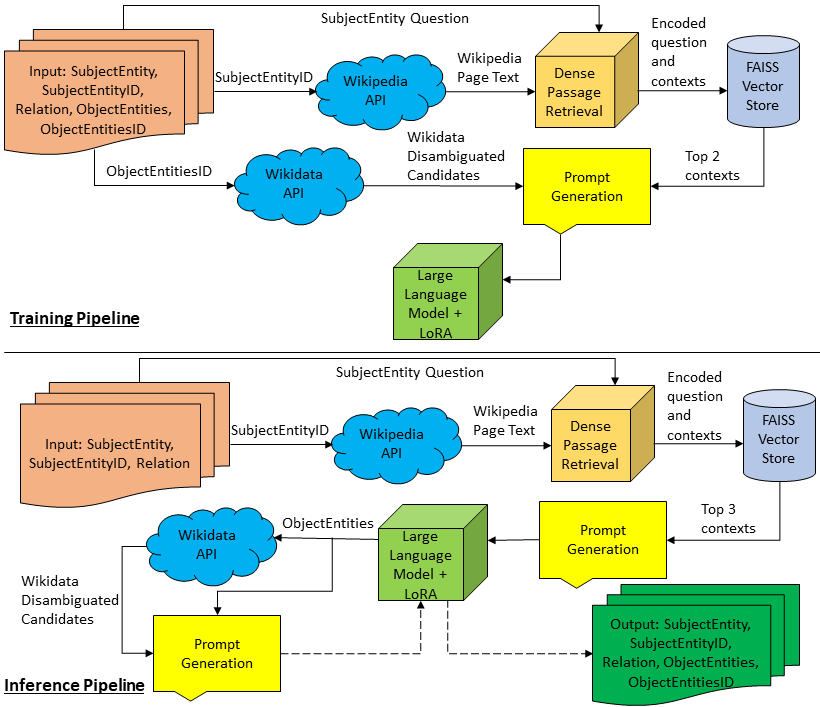}
\caption{LLM2KB System Architecture.} \label{fig1}
\end{figure}

\section{LLM2KB System Description}
\subsection{Components}
\subsubsection{Large Language Model:}We have selected 'Llama-2-13b-chat' model from Meta AI and 'StableBeluga-13B' model from Stability AI as our base models, as they have been fine-tuned for instruction following, are open source and have achieved state-of-the-art performance on numerous benchmarks. Both the models are fine-tuned versions of the original Llama-2-13b base model and have 13 billion parameters. Llama-2-13b-chat is fine-tuned for chat instructions using Supervised Fine Tuning (SFT) and Reinforcement Learning with Human Feedback (RLHF). StableBeluga-13B is a Llama-2-13b model fine-tuned on an Orca~\cite{23} style instruction dataset.

\subsubsection{Context Retrieval Model:}We have selected Dense Passage Retrieval (DPR) from Meta AI~\cite{24} for subject entity context retrieval. DPR comprises a collection of tools and models used in state-of-the-art open-domain Question and Answer research. We used 'dpr-ctx\_encoder-single-nq-base' for encoding questions and 'dpr-ctx\_encoder-multiset-base' for encoding contexts. Further, we store the dense vectors of the contexts in a Facebook AI Similarity Search (FAISS) vector store for fast search and retrieval.

\subsubsection{Entity Disambiguation:}Once we have predicted an object entity, we send the object entity text to the Wikidata API to fetch potential disambiguated entities. The results returned by Wikidata are then given to the LLM to pick the correct disambiguated entity.

\subsection{LLM prompts}
\label{prompts}
We created 4 different prompts based on the format of each of the base models, where Prompt 1,2 and 3 are used for performing instruction tuning and Prompt 4 is used only during inference for in-context learning. \textbf{Note: During inference, Prompt 1,2 and 3 are trimmed after [/INST] for Llama-2-13b-chat and after \#\#\# Assistant\textbackslash nAnswer: for StableBeluga-13B.} Instruction tuning is a technique to perform supervised fine tuning of models to make them learn to produce valid responses for specific instructions. In our case, the instruction tuning helps the model in 2 ways:

\begin{enumerate}
    \item Learn to answer relevant object entities for a given question (see Table~\ref{tab1}) with and without context. This is achieved with Prompt 1 and Prompt 2.
    \item Given a predicted surface string of an object entity, learn to pick the correct Wikipedia entity title (but not Wikidata ID) from a list of candidate options. This is achieved with Prompt 3.
\end{enumerate}

In-context learning is a method where few examples of the expected output are supplied in the prompt and the model uses them to produce an output in a similar fashion. Prompt 1,2 and 3 are expecting the LLM to produce an answer in a Python list of string format, however we noticed that the model sometimes produces a syntactically incorrect list. Hence, we use Prompt 4 to demonstrate to the LLM an incorrect format Answer vs. correct format Answer.\\

\subsubsection{Llama-2-13b-chat prompts:}

\begin{enumerate}
    \item \textbf{\textless s\textgreater[INST] \textless\textless SYS\textgreater\textgreater\textbackslash n
 You are a helpful, respectful and honest assistant. Your answers should be crisp, short and not repititive.\textbackslash n Give valid wikipedia page titles in the answer. The answer should be in a python list of string format.\textbackslash n If you dont know the answer from both the given context and your past knowledge, answer should just be a python empty list.\textbackslash n\textbackslash n \textless\textless/SYS\textgreater\textgreater\textbackslash n context: '\{context\}'\textbackslash n\textbackslash n \{question\} [/INST] Answer: \{answer\} \textless /s\textgreater}

In this prompt, the \{question\} variable is formed by first picking the relevant question corresponding to the relation (see Table~\ref{tab1}) and then replacing the subject entity of the data row in the \{question\}. \{context\} variable is formed by concatenating the strings of the top 2 contexts returned by the DPR system for the given \{question\}. \textbf{It is important to note that to fetch the top 2 contexts, the DPR context encoder is only fed the Wikipedia page textual content of the subject entity and we do not use the content of the Wikipedia Infobox (found at the top right corner of a Wikipedia page), since the Infobox already contains semi-structured information about an entity which would defeat the purpose of using a LLM.} The \{answer\} variable is formed by replacing the object entities of the data row.

    \item \textbf{\textless s\textgreater[INST] \textless\textless SYS\textgreater\textgreater\textbackslash n
 You are a helpful, respectful and honest assistant. Your answers should be crisp, short and not repititive.\textbackslash n Give valid wikipedia page titles in the answer. The answer should be in a python list of string format.\textbackslash n If you dont know the answer from both the given context and your past knowledge, answer should just be a python empty list.\textbackslash n\textbackslash n \textless\textless/SYS\textgreater\textgreater\textbackslash n context: ''\textbackslash n\textbackslash n \{question\} [/INST] Answer: \{answer\} \textless /s\textgreater}

 In this prompt, the \{question\} variable and \{answer\} variable are formed in the same method as the previous prompt, however we do not supply any context.
    \item \textbf{\textless s\textgreater[INST] \textless\textless SYS\textgreater\textgreater\textbackslash n
 You are a helpful, respectful and honest assistant. Your answers should be crisp, short and not repititive.\textbackslash n Choose an answer from the options in the context.\textbackslash n If you dont know the answer from the given context, answer should just be a python empty list.\textbackslash n \textless\textless/SYS\textgreater\textgreater\textbackslash n context: '\{options\}'\textbackslash n\textbackslash n \{question\} [/INST] Answer: \{answer\} \textless /s\textgreater}

 In this prompt, the \{question\} variable is formed in the same method as the previous prompts, the \{options\} variable is the list of titles for the Wikidata entities returned by the API when queried with a given object entity text. Each separate object entity will lead to 1 unique training sample. For e.g. if a data row has 4 object entities, then we will generate 4 separate training samples for it. The \{answer\} variable is the title of the response entity which has the same Wikidata ID as the object entity.

    \item \textbf{\textless s\textgreater[INST] \textless\textless SYS\textgreater\textgreater\textbackslash n
Example 1: Wrong Format: ['People's Republic of China', 'Laos', 'Thailand', 'India', 'Bangladesh']"].} 

\textbf{Correct Format: Answer: "People's Republic of China", "Laos",}

\textbf{"Thailand", "India", "Bangladesh"] \textless/s\textgreater\textbackslash n}

\textbf{Example 2: Wrong Format: ['Artibonite', 'Nord-Est Department', 'South Department','West Department', 'Centre Department',} 

\textbf{ 'Grand'Anse Department', 'North Department']. Correct Format: Answer: "Artibonite", "Nord-Est Department", "South Department", "West Department", "Centre Department", "Grand'Anse Department", "North Department"] \textless/s\textgreater\textbackslash n}

\textbf{Example 3: Wrong Format: ['book's and page's']. Correct Format: Answer: ["book's and page's"] \textless/s\textgreater\textbackslash n
Your answer should only be a valid python list of string format. Do not give any explainations.\textbackslash n \textless\textless/SYS\textgreater\textgreater\textbackslash n Use the examples to convert \{answer\} into a correct python list. [/INST] Answer:}

Once an answer is generated from Prompt 1, 2 or 3, it is used as the \{answer\} variable in this prompt to format the answer in a correct Python list of string format.\\\\\\
\end{enumerate}

\begin{table}
\caption{Questions corresponding to each relation.}\label{tab1}
\begin{tabular}{|l|l|}
\hline
\textbf{Relation} &  \textbf{Question}\\
\hline
BandHasMember	& Who are the members of \_?\\
CityLocatedAtRiver	&Which river is \_ located at?\\
CompanyHasParentOrganisation&	What is the parent organization of \_?\\
CompoundHasParts&	What are the components of \_?\\
CountryBordersCountry	&Which countries border \_?\\
CountryHasOfficialLanguage	&What is the official language of \_?\\
CountryHasStates&	Which states are part of \_?\\
FootballerPlaysPosition	&What position does \_ play in football?\\
PersonCauseOfDeath	&What caused the death of \_?\\
PersonHasAutobiography&	What is the title of \_'s autobiography?\\
PersonHasEmployer&	Who is \_'s employer?\\
PersonHasNoblePrize	&In which field did \_ receive the Nobel Prize?\\
PersonHasNumberOfChildren&	How many children does \_ have?\\
PersonHasPlaceOfDeath&	Where did \_ die?\\
PersonHasProfession&	What is \_'s profession?\\
PersonHasSpouse&	Who is \_ married to?\\
PersonPlaysInstrument&	What instrument does \_ play?\\
PersonSpeaksLanguage&	What languages does \_ speak?\\
RiverBasinsCountry	&In which country can you find the \_ river basin?\\
SeriesHasNumberOfEpisodes	&How many episodes does the series \_ have?\\
StateBordersState&	Which states border the state of \_?\\
\hline
\end{tabular}
\end{table}

\subsubsection{StableBeluga-13B prompts:}

\begin{enumerate}
    \item \textbf{\#\#\# System:\textbackslash nYou are a helpful, respectful and honest assistant. Your answers should be crisp, short and not repititive.\textbackslash nGive valid wikipedia page titles in the answer. The answer should be in a python list of string format.\textbackslash nIf you dont know the answer from both the given context and your past knowledge, answer should just be a python empty list.\textbackslash n\textbackslash n\#\#\# User:\textbackslash ncontext: \{context\}\textbackslash n\textbackslash n\{question\}\textbackslash n\textbackslash n\#\#\# Assistant\textbackslash nAnswer: \{answer\}}

    \item \textbf{\#\#\# System:\textbackslash nYou are a helpful, respectful and honest assistant. Your answers should be crisp, short and not repititive.\textbackslash nGive valid wikipedia page titles in the answer. The answer should be in a python list of string format.\textbackslash nIf you dont know the answer from both the given context and your past knowledge, answer should just be a python empty list.\textbackslash n\textbackslash n\#\#\# User:\textbackslash ncontext: ''\textbackslash n\textbackslash n\{question\}\textbackslash n\textbackslash n\#\#\# Assistant\textbackslash nAnswer: \{answer\}}

    \item \textbf{\#\#\# System:\textbackslash nYou are a helpful, respectful and honest assistant. Your answers should be crisp, short and not repititive.\textbackslash nChoose an answer from the options in the context.\textbackslash nIf you dont know the answer from both the given context, answer should just be a python empty list.\textbackslash n\textbackslash n\#\#\# User:\textbackslash ncontext: \{options\}\textbackslash n\textbackslash n\{question\}\textbackslash n}
    
    \textbf{\textbackslash n\#\#\# Assistant\textbackslash nAnswer: ['\{answer\}']}

    \item \textbf{\#\#\# System:\textbackslash n
    Example 1: Wrong Format: ['People's Republic of China', 'Laos', 'Thailand', 'India', 'Bangladesh']"].} 
    
    \textbf{Correct Format: Answer: "People's Republic of China", "Laos",}
    
    \textbf{"Thailand", "India", "Bangladesh"] \textless/s\textgreater\textbackslash n}
    
    \textbf{Example 2: Wrong Format: ['Artibonite', 'Nord-Est Department', 'South Department','West Department', 'Centre Department',} 
    
    \textbf{ 'Grand'Anse Department', 'North Department']. Correct Format: Answer: "Artibonite", "Nord-Est Department", "South Department", "West Department", "Centre Department", "Grand'Anse Department", "North Department"] \textless/s\textgreater\textbackslash n}
    
    \textbf{Example 3: Wrong Format: ['book's and page's']. Correct Format: Answer: ["book's and page's"]\textbackslash n\textbackslash n}

    \textbf{\#\#\# User:\textbackslash n
    Your answer should only be a valid python list of string format. Do not give any explainations.\textbackslash n\textbackslash n Use the examples to convert \{answer\} into a correct python list.\textbackslash n\textbackslash n \#\#\#Assistant\textbackslash nAnswer:}
\end{enumerate}

The \{question\}, \{answer\}, \{context\} and \{options\} variables are generated in the same method as described for the Llama-2-13b-chat prompts.

\subsection{Training}
For both base models namely Llama-2-13b-chat and StableBeluga-13B, we load them in 4 bit quantized state with frozen weights and only train an injection model using LoRA technique. The injection model has $\approx$0.05 \% of the parameters of the base models. The training setup was as follows:

\begin{itemize}
    \item \textbf{Libraries}: BitsandBytes~\cite{bnb1,bnb2}, HuggingFace~\cite{hf}, PyTorch~\cite{torch}
    \item \textbf{Base model}: BitsandBytesConfig(load\_in\_4bit=True,
    
    bnb\_4bit\_use\_double\_quant=True,

    bnb\_4bit\_quant\_type="nf4", bnb\_4bit\_compute\_dtype=torch.bfloat16)
    \item \textbf{LoRA model}: alpha=16, dropout=0.05, r=4, bias="none", 
    
    task\_type="CAUSAL\_LM"
    \item \textbf{Trainer}: epochs=3, optimizer="paged\_adamw\_32bit",
    
    gradient\_accumulation\_steps=2, per\_device\_train\_batch\_size=1,
    
    per\_device\_eval\_batch\_size= 4, fp16=True,
    
    learning\_rate=2e-5, max\_grad\_norm=0.3, warmup\_ratio=0.03,
    
    lr\_scheduler\_type="constant"

    \item \textbf{GPU}: 2x NVIDIA V100
\end{itemize}

\subsection{Inference} The overall architecture can be seen in Fig.~\ref{fig1}. During inference, we first try to fetch the English language Wikipedia page text of the subject entity using its Wikidata ID. If the subject entity does not have an English page, we then pick the text from the primary alternate language page. We then split the text into chunks of 300 tokens (with an overlap of 50 tokens to maintain continuity) due to the following reasons:
\begin{itemize}
    \item LLM have a limit on the context length, which in the case of Llama-2-13b-chat and StableBeluga-13B is 4096 tokens. Since the Wikipedia text is usually large and may cross the maximum context length supported by the LLM, it is not feasible to consume the entire context in one shot.
    \item Due to the size of the LLM, passing lengthy inputs increases the inference time. Further, it makes much more sense to pick those chunks of information from the entire Wikipedia page that are relevant to a given question.
\end{itemize}

Each of the context chunks are encoded using the DPR context encoder and stored in a FAISS vector store for fast search and retrieval. To ensure that the LLM is not just fixated on using only top 2 contexts, during inference we pick top 3 contexts to test its robustness towards handling variability. To pick the top 3 relevant context chunks for a given question, the question is encoded using the DPR question encoder and then passed to FAISS. The top 3 retrieved context chunks are concatenated to form the \{context\} variable for the prompts. In cases where a subject entity does not have a Wikipedia page, the \{context\} variable will be empty for the prompt, and the LLM will have to rely on its stored knowledge to answer the question. Prompt 1 is executed if a context was found whereas Prompt 2 is executed if no context was found.

Once the LLM processes the prompt, the answer can have 0 or more object entities. For disambiguating each of these surface strings, we query the Wikidata API with each of these object entities separately. The API will return a list of candidate Wikidata entities, from which we attempt to find the correct disambiguated entity. To do this, we collect all the candidate entities and put them in a list to form the \{context\} in Prompt 3. The expectation is that the LLM picks the correct disambiguated entity relevant to the question. For e.g. if the question is "Who is William Shakespeare married to?", and the LLM correctly answers "Anne Hathaway", Wikidata API will return entities of both Anne Hathaway (wife of William Shakespeare) and Anne Hathaway (American actress). With Prompt 3, we believe that providing the question to the LLM will help us to disambiguate and pick the correct candidate. The Wikidata ID of each disambiguated object entity is then fetched and stored. In cases where the LLM generated a string which was not an exact match of the provided options, we pass the LLM output to Wikipedia API for fetching the most relevant entity.

\begin{table}
\caption{LLM2KB results (Precision, Recall and F-1 score are macro average per relation) on Test set with Llama-2-13b-chat. Train + Validation set were used to generate training samples.}\label{tab2}
\begin{tabular}{|l|l|l|l|}
\hline
\textbf{Relation} &  \textbf{Precision}&  \textbf{Recall}&  \textbf{F1 score}\\
\hline
BandHasMember	&0.7178 &0.4151 &0.4857\\
CityLocatedAtRiver	&0.7400 &0.5205 &0.5586\\
CompanyHasParentOrganisation&0.7200 &0.7200 &0.6400\\
CompoundHasParts&0.9318 &0.8495 &0.8784\\
CountryBordersCountry& 0.8492& 0.5848&0.6719\\
CountryHasOfficialLanguage&0.9154 & 0.7500&0.8015\\
CountryHasStates&0.5687 &0.3872 &0.4399\\
FootballerPlaysPosition	&0.7050 &0.6683 &0.6783\\
PersonCauseOfDeath	&0.8500 &0.8200 &0.8200\\
PersonHasAutobiography& 0.6700& 0.4075&0.4173\\
PersonHasEmployer&0.5100 &0.2957 &0.3327\\
PersonHasNoblePrize	&0.9900 &0.9350 &0.9367\\
PersonHasNumberOfChildren&0.4900 &0.4900 &0.4900\\
PersonHasPlaceOfDeath&0.8300 & 0.7600&0.7200\\
PersonHasProfession&0.6600 &0.4628 &0.5034\\
PersonHasSpouse&0.7700 &0.5850 &0.5867\\
PersonPlaysInstrument&0.7633 &0.6159 &0.6354\\
PersonSpeaksLanguage&0.8500 &0.7538 &0.7704\\
RiverBasinsCountry	& 0.8883&0.7689 &0.7896\\
SeriesHasNumberOfEpisodes&0.4500 &0.4500 &0.4500\\
StateBordersState&0.5090 &0.3416 &0.3815\\
\hline
\textbf{Average}&\textbf{0.7323} &\textbf{0.5991} &\textbf{0.6185}\\
\hline
\end{tabular}
\end{table}

\begin{table}
\caption{LLM2KB results (Precision, Recall and F-1 score are macro average per relation) on Test set with Llama-2-13b-chat. Only Train set was used to generate training samples.}\label{tab3}
\begin{tabular}{|l|l|l|l|}
\hline
\textbf{Relation} &  \textbf{Precision}&  \textbf{Recall}&  \textbf{F1 score}\\
\hline
BandHasMember	&0.7235&	0.4132	&0.4822\\
CityLocatedAtRiver	&0.7200	&0.4948	&0.5291\\
CompanyHasParentOrganisation&	0.7400	&0.7550	&0.6667\\
CompoundHasParts&	0.9280	&0.7909	&0.8442\\
CountryBordersCountry	&0.8439&	0.5501&	0.6442\\
CountryHasOfficialLanguage&	0.9000&	0.7372	&0.7877\\
CountryHasStates	&0.5602&	0.3669&	0.4221\\
FootballerPlaysPosition	&0.7450	&0.4900	&0.4917\\
PersonCauseOfDeath&	0.8600&	0.8133	&0.8150\\
PersonHasAutobiography	&0.6900	&0.4125	&0.4240\\
PersonHasEmployer	&0.4100&	0.2237&	0.2555\\
PersonHasNoblePrize&	0.9700&	0.9550&	0.9567\\
PersonHasNumberOfChildren	&0.4300&	0.4300&	0.4300\\
PersonHasPlaceOfDeath&	0.8600&	0.8100&	0.7800\\
PersonHasProfession	&0.6050	&0.4560	&0.4856\\
PersonHasSpouse	&0.7600&	0.5650&	0.5667\\
PersonPlaysInstrument	&0.7100&	0.5982&	0.6133\\
PersonSpeaksLanguage&	0.8650&	0.7327	&0.7607\\
RiverBasinsCountry&	0.8925&	0.7851&	0.7997\\
SeriesHasNumberOfEpisodes&	0.4800&	0.4800&	0.4800\\
StateBordersState	&0.5977&	0.3352	&0.3983\\
\hline
\textbf{Average}&\textbf{0.7281} &\textbf{0.5807} &\textbf{0.6016}\\
\hline
\end{tabular}
\end{table}

\begin{table}
\caption{LLM2KB results (Precision, Recall and F-1 score are macro average per relation) on Test set with StableBeluga-13B. Train + Validation set were used to generate training samples.}\label{tab4}
\begin{tabular}{|l|l|l|l|}
\hline
\textbf{Relation} &  \textbf{Precision}&  \textbf{Recall}&  \textbf{F1 score}\\
\hline
BandHasMember &	0.7075&	0.4227&	0.4850\\
CityLocatedAtRiver	&0.7300	&0.4865&	0.5225\\
CompanyHasParentOrganisation	&0.8800	&0.6500	&0.6300\\
CompoundHasParts&	0.8573&	0.6581&	0.7197\\
CountryBordersCountry	&0.8360	&0.4690&	0.5654\\
CountryHasOfficialLanguage	&0.9462	&0.7372	&0.8021\\
CountryHasStates	&0.7984&	0.3225&	0.3853\\
FootballerPlaysPosition&	0.7100&	0.4933&	0.5050\\
PersonCauseOfDeath&	0.8800&	0.8000&	0.8000\\
PersonHasAutobiography	&0.8100&	0.3450&	0.3533\\
PersonHasEmployer&	0.6700&	0.2128	&0.2403\\
PersonHasNoblePrize	&0.9800&	0.7700&	0.7700\\
PersonHasNumberOfChildren&	0.4400&	0.4400&	0.4400\\
PersonHasPlaceOfDeath&	0.9400&	0.6700	&0.6700\\
PersonHasProfession&	0.6100	&0.3962&	0.4457\\
PersonHasSpouse&	0.8500&	0.4450	&0.4467\\
PersonPlaysInstrument&	0.7667&	0.5577&	0.6055\\
PersonSpeaksLanguage&	0.8850	&0.6190	&0.6947\\
RiverBasinsCountry&	0.9170&	0.7684&	0.8030\\
SeriesHasNumberOfEpisodes&	0.4200&	0.4200	&0.4200\\
StateBordersState	&0.6817&	0.2269	&0.2872\\
\hline
\textbf{Average}&\textbf{0.7769} &\textbf{0.5195} &\textbf{0.5520}\\
\hline
\end{tabular}
\end{table}

\section{Results and Future Work}

The relation level performance of our system can be seen in Table~\ref{tab2}, Table~\ref{tab3} and Table~\ref{tab4}. Our primary experiments involved utilising training samples generated using method 1 for Llama-2-13b-chat and StableBeluga-13B. Additionally, we also trained a Llama-2-13b-chat model with method 2 used to generate training samples. Our best performing model was Llama-2-13b-chat when trained on instruction samples generated using method 1. In training with both method 1 and method 2 data generation strategies, Llama-2-13b-chat performed better than StableBeluga-13B. 

'PersonHasNoblePrize' is the highest scoring relation from our best model and 'PersonHasEmployer' is the lowest scoring relation from our best model. All the 3 models did not perform very highly on the 2 relations which only expected numeric answers namely 'PersonHasNumberOfChildren' and 'SeriesHasNumberOfEpisodes' even after supplying context. After going through the Wikipedia contexts for the subject entities of these relations, we observed that a large majority of them never mention anything about number of children or number of episodes, leaving the LLM to solely rely on its memory of factual knowledge. Overall we observed that the following are some of the practical challenges after subjective analysis of the results:

\begin{itemize}
    \item Fragility of LLM during inference towards minor changes in prompts.
    \item Hallucinations of LLM.
    \item LLM produces a correct surface string of an object but Wikidata API is unable to return any relevant entities for it.
\end{itemize}

Our future work will focus on utilising 30 billion and 70 billion versions of LLM to see if that can push the performance further and also experiment with Chain-of-Thought prompt techniques.


%
%
%

\begin{thebibliography}{8}
\bibitem{gpt}
OpenAI, 2023. GPT-4 Technical Report, 33. arXiv preprint arXiv:2303.08774.

\bibitem{llama2}
Touvron, H., Martin, L., Stone, K., Albert, P., Almahairi, A., Babaei, Y., Bashlykov, N., Batra, S., Bhargava, P., Bhosale, S. and Bikel, D., 2023. Llama 2: Open foundation and fine-tuned chat models. arXiv preprint arXiv:2307.09288.

\bibitem{beluga}
Stability AI Stable Beluga, \url{https://stability.ai/blog/stable-beluga-large-instruction-fine-tuned-models}. Last accessed 7 Aug 2023.

\bibitem{lora}
Hu, E.J., Shen, Y., Wallis, P., Allen-Zhu, Z., Li, Y., Wang, S., Wang, L. and Chen, W., 2021. Lora: Low-rank adaptation of large language models. arXiv preprint arXiv:2106.09685.

\bibitem{lmkbc2023}
LM-KBC Homepage, \url{https://lm-kbc.github.io/challenge2023/}. Last accessed 7 Aug 2023.

\bibitem{petroni}
Petroni, F., Rocktäschel, T., Riedel, S., Lewis, P., Bakhtin, A., Wu, Y. and Miller, A., 2019, November. Language Models as Knowledge Bases?. In Proceedings of the 2019 Conference on Empirical Methods in Natural Language Processing and the 9th International Joint Conference on Natural Language Processing (EMNLP-IJCNLP) (pp. 2463-2473).

\bibitem{lmandfor}
Razniewskia, S., Yatesa, A., Kassnerc, N. and Weikuma, G., 2021. Language Models As or For Knowledge Bases.

\bibitem{knowwhatlmknows}
Jiang, Z., Xu, F.F., Araki, J. and Neubig, G., 2020. How can we know what language models know?. Transactions of the Association for Computational Linguistics, 8, pp.423-438.

\bibitem{inducbert}
Bouraoui, Z., Camacho-Collados, J. and Schockaert, S., 2020, April. Inducing relational knowledge from BERT. In Proceedings of the AAAI Conference on Artificial Intelligence (Vol. 34, No. 05, pp. 7456-7463).

\bibitem{9}
Haviv, A., Berant, J. and Globerson, A., 2021, April. BERTese: Learning to Speak to BERT. In Proceedings of the 16th Conference of the European Chapter of the Association for Computational Linguistics: Main Volume (pp. 3618-3623).

\bibitem{10}
Shin, T., Razeghi, Y., Logan IV, R.L., Wallace, E. and Singh, S., 2020, November. AutoPrompt: Eliciting Knowledge from Language Models with Automatically Generated Prompts. In Proceedings of the 2020 Conference on Empirical Methods in Natural Language Processing (EMNLP) (pp. 4222-4235).

\bibitem{11}
Qin, G. and Eisner, J., 2021, June. Learning How to Ask: Querying LMs with Mixtures of Soft Prompts. In Proceedings of the 2021 Conference of the North American Chapter of the Association for Computational Linguistics: Human Language Technologies (pp. 5203-5212).

\bibitem{12}
Zhong, Z., Friedman, D. and Chen, D., 2021, June. Factual Probing Is [MASK]: Learning vs. Learning to Recall. In Proceedings of the 2021 Conference of the North American Chapter of the Association for Computational Linguistics: Human Language Technologies (pp. 5017-5033).

\bibitem{13}
Fichtel, L., Kalo, J.C. and Balke, W.T., 2021, June. Prompt tuning or fine-tuning-investigating relational knowledge in pre-trained language models. In 3rd Conference on Automated Knowledge Base Construction.

\bibitem{14}
He, T., Cho, K. and Glass, J., 2021. An empirical study on few-shot knowledge probing for pretrained language models. arXiv preprint arXiv:2109.02772.

\bibitem{15}
Kalo, J.C., Fichtel, L., Ehler, P. and Balke, W.T., 2020. KnowlyBERT-Hybrid query answering over language models and knowledge graphs. In The Semantic Web–ISWC 2020: 19th International Semantic Web Conference, Athens, Greece, November 2–6, 2020, Proceedings, Part I 19 (pp. 294-310). Springer International Publishing.

\bibitem{16}
Arnaout, H., Tran, T.K., Stepanova, D., Gad-Elrab, M.H., Razniewski, S. and Weikum, G., 2022. Utilizing Language Model Probes for Knowledge Graph Repair.

\bibitem{17}
Biswas, R., Sofronova, R., Alam, M., Heist, N., Paulheim, H. and Sack, H., 2021. Do judge an entity by its name! entity typing using language models. In The Semantic Web: ESWC 2021 Satellite Events: Virtual Event, June 6–10, 2021, Revised Selected Papers 18 (pp. 65-70). Springer International Publishing.

\bibitem{lmkbc}
Singhania, S., Nguyen, T.P. and Razniewski, S., 2022. LM-KBC: Knowledge Base Construction from Pre-trained Language Models.

\bibitem{18}
Li, T., Huang, W., Papasarantopoulos, N., Vougiouklis, P. and Pan, J.Z., 2022. Task-specific Pre-training and Prompt Decomposition for Knowledge Graph Population with Language Models.

\bibitem{19}
Alivanistos, D., Santamaría, S.B., Cochez, M., Kalo, J.C., van Krieken, E. and Thanapalasingam, T., 2022. Prompting as Probing: Using Language Models for Knowledge Base Construction.

\bibitem{gpt3}
Brown, T., Mann, B., Ryder, N., Subbiah, M., Kaplan, J.D., Dhariwal, P., Neelakantan, A., Shyam, P., Sastry, G., Askell, A. and Agarwal, S., 2020. Language models are few-shot learners. Advances in neural information processing systems, 33, pp.1877-1901.

\bibitem{20}
Ning, X. and Celebi, R., 2022. Knowledge Base Construction from Pre-trained Language Models by Prompt learning.

\bibitem{21}
Fang, X., Kalinowski, A., Zhao, H., You, Z., Zhang, Y. and An, Y., 2022. Prompt Design and Answer Processing for Knowledge Base Construction from Pre-trained Language Models (KBC-LM).
\bibitem{22}
Dalal, S., Sharma, A., Jain, S. and Dave, M., 2022. Manual Prompt Generation For Language Model Probing.
\bibitem{23}
Mukherjee, S., Mitra, A., Jawahar, G., Agarwal, S., Palangi, H. and Awadallah, A., 2023. Orca: Progressive learning from complex explanation traces of gpt-4. arXiv preprint arXiv:2306.02707.

\bibitem{24}
Karpukhin, V., Oguz, B., Min, S., Lewis, P., Wu, L., Edunov, S., Chen, D. and Yih, W.T., 2020, November. Dense Passage Retrieval for Open-Domain Question Answering. In Proceedings of the 2020 Conference on Empirical Methods in Natural Language Processing (EMNLP) (pp. 6769-6781).

\bibitem{bnb1}
Dettmers, T., Lewis, M., Belkada, Y. and Zettlemoyer, L., 2022. Llm. int8 (): 8-bit matrix multiplication for transformers at scale. arXiv preprint arXiv:2208.07339.

\bibitem{bnb2}
Dettmers, T., Lewis, M., Shleifer, S. and Zettlemoyer, L., 2021. 8-bit optimizers via block-wise quantization. arXiv preprint arXiv:2110.02861.

\bibitem{hf}
Wolf, T., Debut, L., Sanh, V., Chaumond, J., Delangue, C., Moi, A., Cistac, P., Rault, T., Louf, R., Funtowicz, M. and Davison, J., 2019. Huggingface's transformers: State-of-the-art natural language processing. arXiv preprint arXiv:1910.03771.

\bibitem{torch}
Paszke, A., Gross, S., Massa, F., Lerer, A., Bradbury, J., Chanan, G., Killeen, T., Lin, Z., Gimelshein, N., Antiga, L. and Desmaison, A., 2019. Pytorch: An imperative style, high-performance deep learning library. Advances in neural information processing systems, 32.

\end{thebibliography}
%

\end{document}